# A General Multiple Data Augmentation Based Framework for Training Deep Neural Networks

Binyan Hu, Yu Sun, and A. K. Qin
*Department of Computing Technologies*
*Swinburne University of Technology*
Melbourne, Australia
Email: bhu@swin.edu.au, yusun@swin.edu.au, kqin@swin.edu.au

*Abstract*—Deep neural networks (DNNs) often rely on massive labelled data for training, which is inaccessible in many applications. Data augmentation (DA) tackles data scarcity by creating new labelled data from available ones. Different DA methods have different mechanisms and therefore using their generated labelled data for DNN training may help improving DNN's generalisation to different degrees. Combining multiple DA methods, namely multi-DA, for DNN training, provides a way to further boost generalisation. Among existing multi-DA based DNN training methods, those relying on knowledge distillation (KD) have received great attention. They leverage knowledge transfer to utilise the labelled data sets created by multiple DA methods instead of directly combining them for training DNNs. However, existing KD-based methods can only utilise certain types of DA methods, incapable of making full use of the advantages of arbitrary DA methods. In this work, we propose a general multi-DA based DNN training framework capable to use arbitrary DA methods. To train a DNN, our framework replicates a certain portion in the latter part of the DNN into multiple copies, leading to multiple DNNs with shared blocks in their former parts and independent blocks in their latter parts. Each of these DNNs is associated with a unique DA and a newly devised loss that allows comprehensively learning from the data generated by all DA methods and the outputs from all DNNs in an online and adaptive way. The overall loss, i.e., the sum of each DNN's loss, is used for training the DNN. Eventually, one of the DNNs with the best validation performance is chosen for inference. We implement the proposed framework by using three distinct DA methods and apply it for training representative DNNs. Experimental results on the popular benchmarks of image classification demonstrate the superiority of our method to several existing single-DA and multi-DA based training methods.

*Keywords — deep neural network, DNN, training, multiple data augmentation, knowledge distillation*

## I. INTRODUCTION

Deep neural networks (DNNs) typically require a massive amount of labelled data for training to achieve desired performance. However, many applications can only access limited data with labels, leading to a high risk of poor generalisation. Data augmentation (DA) aims to tackle data scarcity by creating new labelled data from available ones via certain mechanisms, e.g., randomly cropping and rotating an image. Due to its simplicity and effectiveness, DA has been widely used for training DNNs [1]–[3]. There exist different types of DA methods. For example, according to the ways to dealing with labels when creating new data [4], there exist 1) label-invariant DA (e.g., AutoAugment [5] and RandAugment [6]) which generates new data from some available data while inheriting the labels of the available data, leading to new data with existing labels, and 2) label-mixing DA (e.g., Mixup [7] and CutMix [8]) which mixes not only data but also their labels, leading to new data with new labels.

Different DA methods have different mechanisms of creating new data, yielding data diversity that allows improving generalisation from distinct aspects. To leverage the complementary advantages of different DA methods, it is a common practice to combine multiple DA methods, i.e., multi-DA, for training a DNN to boost its generalisation. There exist different multi-DA based methods for training DNNs. The most intuitive ones are to create new data using multiple DA methods, in a parallel or sequential manner, and then use the created data for training a DNN [1], [3], [5], [6]. In a parallel manner, each DA method is independently applied to create multiple new data sets which are combined into one training set. In a sequential manner, each DA method is sequentially applied one after another to create the training set. Recently, knowledge distillation (KD) [9], as a very popular knowledge transfer strategy, has been used for multi-DA based DNN training. Instead of directly leveraging the data generated by multiple DA methods, KD-featured methods make use of the outputs of the DNNs trained on the data generated by different DA methods via knowledge transfer. By doing so, richer class information encoded by the softmax outputs of DNNs can be learned. Some KD-based methods [10]–[14] create the multi-views of images and train the DNN on different views while pursuing the consistency of DNN's output w.r.t. each view in an online manner via knowledge distillation. Existing multi-DA based DNN training methods that rely on KD have demonstrated their superiority in practice. However, they can only incorporate certain types of DA methods, incapable to make use of arbitrary DA methods. For example, the above-mentioned KD-based methods require the use of label-invariant DA methods to allow formulating the multi-view consistency, which is off limit to label-mixing DA.

To address this limitation, we propose a general multi-DA based DNN training framework capable of utilising arbitrary DA methods. For training a DNN, our framework replicates a certain portion in the latter part of the DNN into multiple copies, leading to multiple DNNs with shared blocks in their former parts and independent blocks in their latter parts. Each of these DNNs is accompanied by a unique DA and a newly devised loss with self and mutual terms to allow it to comprehensively learn from the data generated by all DA methods and the outputs from all DNNs in an online and adaptive manner. The overall loss that sums over each DNN's loss is used for training the DNN. Finally, one of the DNNs involved in the proposed framework with the best validation performance is chosen to be used for inference. Fig. 1 illustrates the proposed framework with three DA methods as an example.

We implement the framework via three DA methods, i.e., RandAugment [6], Mixup [7], and CutMix [8], which have different working mechanisms and are popularly employed for DNN training [1]–[3]. The implemented method is used for training several DNNs such as ResNet and DenseNet on

This work was performed on the OzSTAR national facility at Swinburne University of Technology. The OzSTAR program receives funding in part from the Astronomy National Collaborative Research Infrastructure Strategy (NCRIS) allocation provided by the Australian Government. This work was supported in part by the Australian Research Council (ARC) under Grant No. LP180100114 and DP200102611.

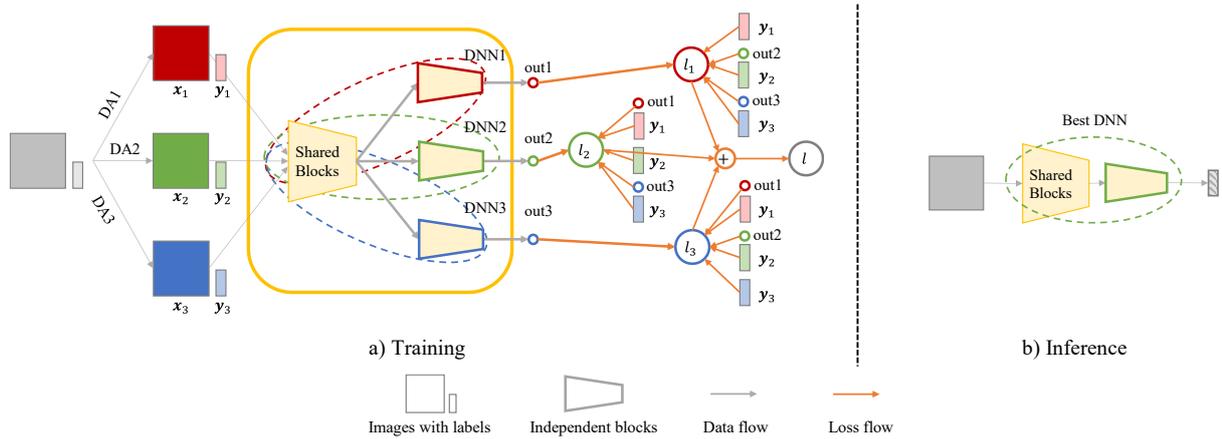

Fig. 1. An illustration of the proposed framework by using three DA methods as an example, i.e., $N = 3$ (best viewed in colour). a) **Training**: (1) Given a DNN to be trained, the framework replicates its latter part into three copies, yielding three DNNs with shared blocks in their former parts and independent blocks in their latter parts. Each of these DNNs is associated with a unique DA. (2) In each step, we feed in each DNN with three images yielded by three DA methods, leading to nine predictions. (3) A loss function $l$, defined as the sum of the newly designed individual loss functions $l_i, i = 1, 2, 3$, for three DNNs, is used for training. $l_i$ is defined on the data from three DA methods and the outputs of three DNNs, as detailed in Section III.B. b) **Inference**: For inference, the best DNN among three DNNs, e.g., DNN2, is selected based on the validation performance, following the procedure detailed in Section III.B. In this figure, three colours are used to differentiate data/model components relevant to different DA methods. $(x_i, y_i), i = 1,2,3$ denotes the image-label pair augmented by the $i$-th DA method, "out$_i$" denotes the output of the $i$-th DNN, and $l_i$ denotes the loss of the $i$-th DNN.

some popular image classification datasets, e.g., CIFAR-10, CIFAR-100, tiny-ImageNet, and ImageNet, and compared to some existing single-DA and multi-DA based training methods. Experimental results demonstrate the superiority of our method. We also perform an ablation study to verify the effectiveness of several key components in the proposed framework.

## II. RELATED WORK

### A. Data Augmentation

Data augmentation (DA) can increase the volume and diversity of data by generating new labelled data from existing ones by applying certain operations, e.g., randomly cropping and rotating an image. It has been widely used for DNN training [1]–[3].

Different DA methods adopt different mechanisms to create new data. Label-invariant DA, such as horizontal flipping, random cropping, random resizing, and colour jittering, generates new data from certain available data with the labels of the available data inherited, leading to new data with existing labels. Recently, the success of AutoML leads to the advent of AutoAugment [5] which can learn to design a DA method by composing multiple label-invariant DA methods. Several follow-up works, like RandAugment [6], aim to improve the generalisation performance of the trained DNN or reduce the computational costs of the design process. Label-mixing DA mixes not only data but also their labels, leading to new data with new labels. Mixup [7] linearly combines the image pixel values and the one-hot labels of any two samples to generate a new labelled sample, as shown in Fig. 2. It can effectively improve the generalisation of the trained model. There exist some more sophisticated label-mixing DA methods like Cut-Mix [8]. Notably, RandAugment, Mixup, and CutMix are among the most popular DA methods for vision tasks. We use them in this work to implement the proposed framework.

To leverage on the complementary merits of different DA methods, it is a common way to combine multiple DA methods, namely multi-DA, to further boost generalisation. There are different multi-DA DNN training methods. The most intuitive ones are to create new data by applying different DA methods in either a parallel or a sequential manner, and then use the created data for training DNNs [1], [3], [5], [6]. The parallel manner independently applies multiple DA methods to create multiple new data sets which are combined into one training set. In a sequential manner, each DA method is sequentially applied one after another to create the training set.

### B. Knowledge Distillation

Knowledge distillation (KD) [9] is one of the most popular transfer learning methods, allowing learning one model to benefit from the processes of learning other models. KD was initially designed for model compression [9], [15], where a small-sized "student" model is trained under the supervision from a large-sized pre-trained "teacher" model. Further developments include: 1) the teacher has a similar capacity to the student [16], [17], 2) the teacher learns from scratch together with the student [18], [19], and 3) the teacher is the student itself [20]. KD may utilise soft labels which better reflect the relationship between different classes, more informative than the ground-truth labels, to help model learning [17], [21]. Also, KD may encourage the output consistency over multiple DNNs, functioning as the implicit regularisation for DNN training to avoid overfitting [18].

KD has been successfully used in DA-based DNN training methods. For example, some works [17], [22], [23] leverage the outputs of a pre-trained model to refine the labels generated a DA method. Recently, some works [10]–[14] make use of online KD (OKD) for multi-DA based DNN training. They generate the multi-views of images and train the DNN on different views subjected to the consistency of DNN's outputs over different views which is imposed via KD. Existing multi-DA based DNN training methods which utilise KD can merely incorporate certain types of DA methods, incapable to make use of arbitrary DA methods. For example, the above-mentioned OKD methods can only use label-invariant DA methods to allow formulating the multi-view consistency, which is off limit to label-mixing DA. In contrast, our proposed framework is capable of incorporating DA methods of arbitrary types.

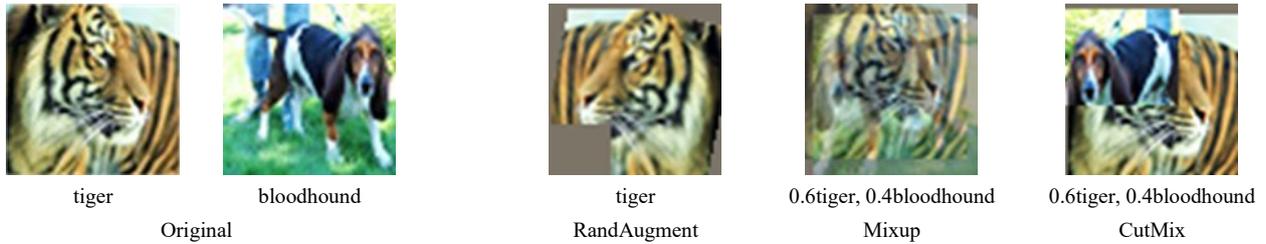

Fig. 2. Example results from three data augmentation methods used in this work, i.e., RandAugment, Mixup, and CutMix, where the original images are taken from the ImageNet dataset.

## III. METHOD

### A. Preliminary

In a typical DNN training process, a DNN, denoted by $f$, learns its parameters $\omega$ from a training set $\mathcal{D}_{train}$, yielding $f^*$ armed with optimised parameters $\omega^*$. The generalisation performance of $f$ is evaluated on a test set $\mathcal{D}_{test}$ which does not overlap with $\mathcal{D}_{train}$. A data augmentation method $T$ aims at creating new data by transforming $\mathcal{D}_{train}$ at certain random so that data volume and diversity may get increased to improve DNN's generalisation. Training $f$ with the help of data augmentation method $T$ can be formulated as

$$\omega^* = \arg\min_{\omega} \left( \mathbb{E}_{(x,y) \sim T(\mathcal{D}_{train})} \mathcal{L}(y, f(x; \omega)) \right) \quad (1)$$

where $(x, y)$ denotes a pair of an image and its one-hot label, $\hat{y} = f(x; \omega)$ denotes model's softmax prediction for $x$, and $\mathcal{L}(y, \hat{y})$ represents the loss function. Our work involves multiple DA methods $\{T_i\}_{i=1}^N$ for DNN training, where $N$ denotes the total number of DA methods. We denote the $T_i$-augmented data sample w.r.t. $(x, y)$ as $(x_i, y_i)$, i.e., $(x_i, y_i) = T_i(x, y)$.

### B. Framework

As shown in Fig. 1, to train a DNN, our framework will replicate a certain portion in the latter part of the network into $N$ copies, leading to $N$ DNNs with shared blocks in their former parts and independent blocks in their latter parts. Each of these DNNs is associated with a unique DA. The reason behind this design is to allow the former part of the network to learn uniform low-level features from the data generated by all DA methods and the latter part of the network to learn distinct high-level features from the data generated by a specific DA. Meanwhile, this design can much save the computational costs for training [14], [19], [24]. Hereafter, we use $f_i$ and $\omega_i$ to represent the $i$-th DNN and its weights, respectively. We use $\hat{y}_i^k$ to denote $f_i$'s softmax prediction result on image $x_k$.

In each training step, we feed in each DNN with $N$ images $\{x_i, i = 1, ..., N\}$ obtained by applying $N$ DA methods to $x$, yielding $N \times N$ predictions $\{\hat{y}_i^k = f_i(x_k), i = 1, ..., N, k = 1, ..., N\}$. The loss function of each DNN is constructed upon these $N \times N$ predictions and $N$ labels $\{y_i, i = 1, ..., N\}$. Specifically, each DNN $f_i$ is subjected to two loss terms: 1) the **self loss** $\ell_i^{sel}$ that relies on $T_i$ associated with $f_i$, 2) the **mutual loss** $\ell_i^{mut}$ which enables $f_i$ to learn from $\{f_k, T_k | k = 1 ... N, k \neq i\}$ through KD. The loss function of $f_i$ is defined as:

$$\ell_i = \ell_i^{sel} + \ell_i^{mut} \quad (2)$$

Herein, the self loss $\ell_i^{sel}$ is calculated on $T_i$-augmented data, i.e.,

$$\ell_i^{sel} = \mathcal{L}(y_i, \hat{y}_i^i) \quad (3)$$

where $\mathcal{L}$ denotes the cross-entropy function. $\ell_i^{sel}$ enables $f_i$ to learn directly from the data augmented by $T_i$. Meanwhile, $f_i$ learns from $\{f_k | k = 1 ... N, k \neq i\}$ via online KD based on mutual loss $\ell_i^{mut}$ which consists of direct loss and KD loss components, where the former is defined via the ground-truth labels of the DA-generated data and the latter is defined on DNN's outputs. Both direct loss and KD loss are defined via cross-entropy, denoted by $\mathcal{L}$. The KD loss aims at providing more comprehensive supervision from the other DNN's outputs.

Training DNNs takes time. At the beginning of the training process, it is unlikely for DNNs trained on DA-generated data to obtain accurate prediction. As a result, the KD loss based supervision may not make effect and even mislead the training, which will be experimentally demonstrated in Section IV.C. To address this issue, by following common practice [9], we define a mutual loss for $f_i$ as a linear combination of the direct loss (w.r.t. ground-truth labels) and the KD loss (w.r.t. other DNNs' outputs):

$$\ell_i^{mut} = \sum_{k=1,...,N, k \neq i} \binom{(1-\beta_k)\mathcal{L}(y_k, \hat{y}_i^k)}{+\beta_k \mathcal{L}(\hat{y}_k^k, \hat{y}_i^k)} \quad (4)$$

where $\beta_k$ is a scalar coefficient to balance the direct and KD loss components. Suppose three DA methods are used to implement the framework, i.e., $N = 3$, this mutual loss term includes two direct and two KD loss components. Taking DNN $f_1$ as an example, as shown in Fig. 1, its two direct loss components are defined on its own outputs and the ground-truth labels w.r.t. the data generated by $T_2$ and $T_3$, respectively. Its two KD loss components are defined on its on its own outputs and the outputs of $f_2$ and $f_3$, w.r.t. the data generated by $T_2$ and $T_3$, respectively.

The overall loss function used for training is defined as:

$$\ell = \sum_{i=1,...,N} \ell_i \quad (5)$$

**$\beta$ scheduler.** It is well known that the KD process relies on good "teacher" DNNs to provide accurate supervision (via soft labels) so that useful knowledge can be transferred to the "student" DNN. In our proposed framework, all of the involved DNNs are trained together from scratch. In the early training phase, DNNs are insufficiently trained and therefore "teacher" DNNs can hardly produce accurate supervision. As a result, the KD loss becomes misleading. It is thus more reasonable to let the direct loss to dominate mutual loss $\ell_i^{mut}$ by setting a small value of $\beta_k$. In the later training phase, DNNs get better trained and therefore the KD loss becomes more informative. Accordingly, the proportion of the KD loss in the mutual loss should be augmented by increasing the value of $\beta_k$.

Following some common practice of parameter adaption [25], we design a β scheduler to adaptively adjust its value. Specifically, instead of using a fixed $\beta_k$, $\beta_k$ is adapted according to the quality of $f_k$'s outputs, measured by "fit" which evaluates how well $f_k$ fits the data generated by $T_k$, as the training proceeds. The "fit" is calculated based on the absolute error between one-hot labels and DNN's softmax outputs. The "fit" for $f_k$ w.r.t. $T_k$-augmented data sample $(x_k, y_k)$ is defined as:

$$fit_k(x_k, y_k) = 1 - \frac{\|y_k - f_k(x_k)\|_1}{2} = 1 - \frac{\|y_k - \hat{y}_k^k\|_1}{2} \quad (6)$$

Given $\|y_k\|_1 = 1$ and $\|\hat{y}_k\|_1 = 1$, based on the triangle inequality, we have $0 \leq \|y_k - \hat{y}_k^k\|_1 \leq \|y_k\|_1 + \|\hat{y}_k^k\|_1 = 2$, where the upper bound is reached when $y_k$ and $\hat{y}_k^k$ completely mismatch. Therefore, the "fit" measure is bounded within [0,1]. The better $y_k$ and $f_k(x_k)$ match, the larger the "fit" value is. During training, every $\beta_k$ is calculated via a momentum-based moving average of $fit_k$, i.e., a linear combination of the old $\beta_k$ value and the newly calculated "fit" value, defined below:

$$\beta_k \leftarrow m\beta_k + (1-m)fit_k(x_k, y_k) \quad (7)$$

where $m \in [0,1)$ is a momentum coefficient and $\beta_k$ is updated after every training batch.

After training a DNN using the proposed framework, one of the DNNs (i.e., shared and independent blocks) involved in the framework, will be selected to be used for inference. We hereby elaborate the selection procedure. First, the whole set of training data is randomly split into 80% for training and 20% for validation. We train the DNN on the training set for multiple times using the proposed framework, calculate the average performance of each DNN involved in the proposed framework on the validation set over multiple training runs, and select the best DNN $i^*$ (involved in the framework) with the best average validation accuracy. Then, we train the DNN on the whole set of training data and choose to use the trained DNN $i^*$ for inference, where the inference performance will be evaluated on a separate test set.

## IV. EXPERIMENTS

**Image Augmentation.** We employ the following three data augmentation methods, as illustrated in Fig. 2, to implement our framework:

- RandAugment [6] is a label-invariant DA method which first randomly selects a fixed number of operations from a set of primitive label-invariant operations like rotation and Cutout [26], and then applies the selected operations in sequence with certain transformation magnitudes, where a specific setting of the magnitude refers to certain parameter settings of operations, e.g., the magnitude of 1 corresponds to the rotation angle of 3° and the cutout size of 4 × 4. There are two crucial parameters in RandAugment, i.e., the number of operations selected every time and the magnitude for applying these operations.
- Mixup [7] is a label-mixing DA method which interpolates both pixel and label values from different samples in a random portion, sampled from a Beta distribution controlled by parameter $\alpha$, i.e., B($\alpha, \alpha$). Mixup is good for improving the robustness to noisy and corrupted samples.
- CutMix [8] is a label-mixing DA method which generates a new image by cutting and pasting two image patches, where the mixing ratio is sampled in the same way as Mixup.

The corresponding label is obtained by interpolating the two labels in proportion to the sizes of the mixed patches. Compared to Mixup, CutMix makes the trained DNN more aware of object locations.

The above three DA methods have been successfully used for DNN training [1]–[3]. They have different nature and thus enable the good diversity of generated data to allow improving DNN generalisation from multiple aspects. We use these three DA methods to validate the effectiveness of our framework without an intention to seek performance optimality.

**Baseline multi-DA based DNN training methods.** We employ two multi-DA based DNN training methods, commonly used for DNN training [1], [3], [5], [6], as baseline approaches to compare with the proposed method. The first one, namely **Baseline 1**, applies a DA method which is uniformly and randomly chosen from the three DA methods used in the proposed method to augment each training sample. The second one, namely **Baseline 2**, first applies RandAugment and then applies one of Mixup and CutMix, chosen uniformly at random, to augment each training sample. We also compare the performance of the proposed method with those obtained via single-DA based training methods w.r.t. each of the three DA methods. Notably, choosing the best one from the three single-DA based methods, denoted by **Best Single DA**, is regarded as a multi-DA based method.

**Common experimental setup.** All DNNs are trained via the SGD optimiser with Nesterov momentum of 0.9 and a cosine annealing learning rate with its initial value set separately for different test cases. The beta distribution parameter $\alpha$ for Mixup and CutMix are both set to 1. The proposed method consumes training data in every training step three times more than the other methods. For a fair comparison, we reduce the number of training steps to 1/3 for the proposed method to ensure that all compared methods consume the same number of training data. The momentum $m$ in the $\beta$ scheduler is set to 0.9.

### A. CIFAR and Tiny-ImageNet Classification

**Datasets.** We first evaluate the proposed method on three small-scale image classification benchmarks: 1) CIFAR-10 [27] with 50K/10K 32×32 images from 10 classes for training/testing. 2) CIFAR-100 [27] with 50K/10K 32×32 images from 100 classes for training/testing. 3) tiny-ImageNet[1] with 100K/10K 64×64 images from 200 classes for training/testing.

**Models.** On CIFAR datasets, we test DNNs with different architectures, incl. ResNet [28], WideResNet (WRN) [29], DenseNet [30], [31], and ResNeXt [32]. On tiny-ImageNet, we use PreAct-ResNet [33]. All employed DNN architectures consist of one stem followed by three blocks and one classification head. In the proposed method, we specify the stem and the first block as shared blocks and the other two blocks and the classification head as independent blocks.

**Setup.** We set the initial learning rate as 0.1. The values of weight decay are set to $5 \times 10^{-4}$ and $10^{-4}$ on CIFAR and tiny-ImageNet, respectively. In all experiments, we specify the number of operations as 1 and the magnitude as 6 for RandAugment. We train ResNet and WRN for 600 epochs, DenseNet and ResNeXt for 400 epochs, and PreAct-ResNet for 200 epochs.

---
[1] https://tiny-imagenet.herokuapp.com/

TABLE I. COMPARISION OF THE ERROR RATES (MEAN ± STD%), AVERAGED OVER THREE RUNS, OBTAINED BY OUR PROPOSED METHOD AND EACH COMPARED METHOD FOR DIFFERENT DNNS ON CIFAR-10, CIFAR-100, AND TINY-IMAGENET DATASETS. EACH ROW MAKES A COMPARISON W.R.T. A SPECIFIC DNN AND A SPECIFIC DATASET, WEHRE THE BEST ERROR RATE (IN TERMS OF THE MEAN VALUE) IS HIGHLIGHTED IN BOLD, AND THE SECOND BEST IS UNDERLINED.

| Dataset | Model | No DA | Single-DA | | | Multi-DA | | | |
|---|---|---|---|---|---|---|---|---|---|
| | | | *RandAug* | *Mixup* | *CutMix* | *Best Single DA* | *Baseline 1* | *Baseline 2* | *Ours* |
| CIFAR10 | ResNet32 | 5.6 ± 0.1 | 4.6 ± 0.0 | 5.6 ± 0.0 | 4.8 ± 0.2 | 4.6 ± 0.0 | <u>4.5 ± 0.1</u> | 4.8 ± 0.1 | **4.4 ± 0.1** |
| | WRN28-2 | 4.6 ± 0.1 | 3.4 ± 0.1 | 4.3 ± 0.1 | 3.8 ± 0.1 | 3.4 ± 0.1 | <u>3.3 ± 0.0</u> | 3.3 ± 0.1 | **3.1 ± 0.1** |
| | WRN40-2 | 4.6 ± 0.0 | 3.0 ± 0.1 | 3.9 ± 0.2 | 3.4 ± 0.1 | 3.0 ± 0.1 | <u>2.9 ± 0.1</u> | 3.1 ± 0.1 | **2.7 ± 0.1** |
| | DenseNet-40 | 5.5 ± 0.0 | 4.3 ± 0.1 | 5.3 ± 0.2 | 4.8 ± 0.1 | <u>4.3 ± 0.1</u> | 4.5 ± 0.1 | 4.7 ± 0.1 | **4.1 ± 0.1** |
| | ResNeXt-29 | 4.4 ± 0.2 | 3.4 ± 0.1 | 3.8 ± 0.0 | 3.6 ± 0.1 | 3.4 ± 0.1 | <u>3.1 ± 0.1</u> | 3.2 ± 0.1 | **3.0 ± 0.1** |
| CIFAR100 | ResNet32 | 28.5 ± 0.0 | 24.9 ± 0.3 | 26.0 ± 0.1 | 26.1 ± 0.3 | <u>24.9 ± 0.3</u> | 25.0 ± 0.1 | 26.4 ± 0.1 | **23.6 ± 0.1** |
| | WRN28-2 | 24.2 ± 0.2 | 20.5 ± 0.2 | 22.1 ± 0.2 | 20.7 ± 0.2 | 20.5 ± 0.2 | <u>19.8 ± 0.3</u> | 20.2 ± 0.3 | **18.4 ± 0.2** |
| | WRN40-2 | 22.6 ± 0.2 | 19.8 ± 0.3 | 20.7 ± 0.1 | 19.9 ± 0.1 | 19.8 ± 0.3 | <u>18.4 ± 0.1</u> | 18.9 ± 0.2 | **17.2 ± 0.1** |
| | DenseNet-40 | 27.1 ± 0.1 | 24.3 ± 0.3 | 25.5 ± 0.1 | 24.9 ± 0.2 | 24.3 ± 0.3 | <u>24.0 ± 0.2</u> | 25.7 ± 0.2 | **22.8 ± 0.2** |
| | ResNeXt-29 (2x32d) | 21.2 ± 0.1 | 18.6 ± 0.3 | 20.0 ± 0.2 | 19.4 ± 0.3 | 18.6 ± 0.3 | <u>17.8 ± 0.3</u> | 18.0 ± 0.3 | **16.8 ± 0.2** |
| tiny-ImageNet | PreAct-ResNet18 | 37.0 ± 0.2 | 34.6 ± 0.2 | 35.1 ± 0.5 | 33.5 ± 0.3 | 33.5 ± 0.3 | 32.1 ± 0.5 | <u>31.7 ± 0.2</u> | **29.6 ± 0.2** |

**Results.** As reported in TABLE I, although Baseline 1 is competitive, its performance does not constantly outperform single-DA based methods over all cases. Especially for small models such as ResNet-32 and DenseNet-40, Baseline 1 hardly makes improvement over the Best Single DA method. Baseline 2 often results in poor performance because applying multiple transformations sequentially to every sample tends to shift the data distribution far from the original. Our method constantly outperforms all of the compared methods and improves on the second best performer by a non-trivial margin in most cases.

*B. ImageNet Classification*

**Dataset.** The ImageNet dataset [34] contains 1.28M/50K images from 1000 classes for training/test. In addition to the original test set, we evaluate the generalisation ability of the trained model on the ImageNet-C test set [35], which is built by corrupting the original test data with operations under four categories including noise, blur, weather, and digital effects at five severity levels. We select 13 corruption operations and use the five corrupted test sets that correspond to the five severity levels for each of them to conduct experiments, i.e., 65 test sets in total.

**Model.** We train ResNet-RS-50 [2], one of the state-of-the-art ResNet variants, which modifies ResNet by incorporating ResNet-D shortcut [36], Squeeze and Excitation [37], etc. The model architecture consists of one stem followed by four blocks and one classification head. Dropout [38] with the probability of 0.25 is adopted before the last fully connected layer in the head for all of the compared methods except our method. This is because Dropout affects model outputs during training and thus may adversely influence the KD process which relies on accurate model outputs. In our method, we define the stem and the first three block as shared blocks and the fourth block and the classification head as independent blocks.

**Setup.** According to [2], [3], we train ResNet-RS-50 for 350 epochs with 5 warm-up epochs, weight decay of $4 \times 10^{-5}$, initial learning rate of 0.4, batch size of 1024, and label smoothing rate of 0.1. The number of operations and the magnitude for RandAugment are set to 2 and 10, respectively. Following the training/inference protocol commonly used for ImageNet dataset, we use randomly cropped image patches with size of 160×160 for training and calculate the top-1 accuracies of the trained model on two sets of centre cropped patches with size of 160×160 and 224×224, respectively. We also test the trained model on ImageNet-C for image patches with size of 224×224.

**Results.** We report the top-1 accuracies on the ImageNet test sets in TABLE II. It can be observed that our method outperforms all compared methods for both of image patch sizes under exam. In the 160×160 patch size case, our method improves the second best performer, i.e., Baseline 1, by 0.64%. In the 224×224 patch size case, our method improves the second by 0.45% while the second surpasses the Best Single DA baseline by 0.74%. The results demonstrate the superiority of the proposed method on large-scale image recognition tasks.

The above-trained model is also evaluated on the 65 test sets from ImageNet-C according to the protocol [35], where a normalisation is applied to each corruption operation. Specifically, for a certain corruption operation, the error rates obtained on its associated five test sets are averaged and divided by the corresponding mean error rate obtained by AlexNet as reported in [35], yielding the corruption-specific normalised mean error rate. A mean corruption error (mCE) is calculated by averaging corruption-specific normalised mean error rates over all corruption operations. Such obtained results are reported in TABLE III. It can be observed that our method outperforms all compared methods in terms of mCE. Mixup ranks second, better than the two baseline multi-DA based methods, revealing their limit in this testing scenario. Notably, our method improves on Mixup by synergising other DA methods and ranks either the first or the second for most corruption operations.

*C. Ablation Study*

We conduct an ablation study to verify the effectiveness of two major components of the proposed framework: 1) the mutual loss configuration and 2) the $\beta$ scheduler.

TABLE II. TOP-1 ACCURACIES (%) OF RESNET-RS-50 ON IMAGENET. ALL MODELS ARE TRAINED WITH 160×160 IMAGE PATCHES AND TESTED FOR IMAGE PATCHES WITH SIZES OF 160×160 AND 224×224, RESPECTIVELY. THE EXPERIMENT IS ONLY RUN ONCE CONSIDERING THE COMPUTATIONAL COSTS. FOR EACH TEST CASE, THE BEST RESULT IS SHOWN IN BOLD, AND THE SECOND BEST IS UNDERLINED.

| Inference patch size | No DA | Single-DA | | | Multi-DA | | | |
|---|---|---|---|---|---|---|---|---|
| | | RandAugment | Mixup | CutMix | Best Single DA | Baseline 1 | Baseline 2 | Ours |
| 160×160 | 76.39 | 78.68 | 78.76 | 78.37 | 78.76 | 79.69 | 79.24 | **80.33** |
| 224×224 | 78.64 | 80.45 | 80.40 | 79.71 | 80.45 | 81.19 | 80.45 | **81.64** |

TABLE III. MEAN CORRUPTION ERRORS (%), NAMELY MCE, AND CORRUPTION-SPECIFIC NORMALISED MEAN ERROR RATES (%) W.R.T. 13 CORRUPTION OPERATIONS OBTAINED BY DIFFERENT COMPARED METHODS ON IMAGENET-C. IMAGE PATCH SIZE OF 224×224 IS USED FOR INFERENCE. FOR EACH CORRUPTION, THE BEST RESULT ACHIEVED AMONG ALL COMPARED METHODS IS SHOWN IN BOLD, AND THE SECOND BEST IS UNDERLINED.

| DA Method | mCE | noise | | | blur | | | | weather | | | digital | | |
|---|---|---|---|---|---|---|---|---|---|---|---|---|---|---|
| | | gaussian | shot | impulse | defocus | glass | motion | zoom | snow | frost | fog | elastic | pixelate | jpeg |
| RandAugment | 67.1 | 65.6 | 65.6 | 63.7 | 75.5 | 91.0 | 74.4 | 75.6 | 62.9 | 67.0 | 41.9 | **78.2** | 48.3 | 62.8 |
| Mixup | 64.0 | 59.1 | 65.5 | 59.8 | 76.6 | **89.2** | 73.8 | 70.1 | 57.5 | 48.2 | 36.5 | 80.5 | 48.2 | 67.3 |
| CutMix | 76.1 | 76.2 | 77.7 | 81.8 | 81.1 | 92.5 | 84.6 | 77.3 | 75.3 | 74.8 | 54.4 | 85.3 | 53.9 | 74.3 |
| Baseline 1 | 64.3 | **58.1** | 60.4 | **57.4** | **73.1** | 90.8 | 74.0 | 70.2 | 59.1 | 60.2 | 40.4 | 79.3 | 49.4 | 63.0 |
| Baseline 2 | 66.5 | 62.4 | 63.2 | 63.2 | 78.4 | 93.1 | 73.8 | 74.1 | **56.9** | 59.7 | 39.3 | 83.6 | 53.1 | 63.9 |
| Ours | **63.0** | 58.2 | **59.8** | 56.9 | 73.2 | 89.9 | **70.2** | **68.9** | 58.8 | **57.3** | **37.7** | 79.0 | **47.9** | **61.4** |

**Effectiveness of the mutual loss configuration.** As elaborated in Section III.B, the mutual loss for every DNN consists of direct loss and KD loss components. We investigate the effectiveness of this design through the following studies. We create three variants of the proposed method in terms of the mutual loss design. The first *"w/o direct loss"* makes the mutual loss merely composed of the KD loss. The second *"w/o KD loss"* makes the mutual loss only composed of the direct loss. The third *"w/o direct and KD losses"* makes the training purely based on the self loss. We choose WRN28-2 and train it on CIFAR-100. The results are reported in TABLE IV. We can observe that once direct or KD loss is not used, the performance degrades drastically. The worst performance is obtained when both direct and KD losses are abandoned, leading to the least amount of knowledge transfer via the self loss and the shared blocks. In contrast, the contributions from direct and KD losses to the overall performance are similar, evidenced by the similar errors when either of them is unused. These results reveal that the mutual loss is a must, where both of its direct and KD loss components are indispensable.

TABLE IV. ABLATION STUDY ON THE MUTUAL LOSS. EACH EXPERIMENT IS RUN THREE TIMES AND AVEAGE ERROR RATES (%) ARE REPORTED. THE BEST VALUE IS SHOWN IN BOLD AND THE SECOND BEST IS UNDERLINED.

| Variant | Error Rate |
|---|---|
| Ours (w/ direct and KD losses) | **18.4** |
| w/o direct loss (all $\beta = 0$) | 19.7 |
| w/o KD loss (all $\beta = 1$) | 19.9 |
| w/o direct and KD losses | 21.9 |

**Effectiveness of the $\beta$ scheduler.** We compare the proposed $\beta$ scheduler with two $\beta$ scheduling schemes: 1) setting $\beta$ to a certain fixed value chosen from [0,1] and 2) linearly increasing $\beta$ from 0 to 1 during training, where $\beta = 0$ and $\beta = 1$ correspond to *"w/o direct loss"* and *"w/o KD loss"*, respectively. We choose WRN28-2 and train it on CIFAR-100 to compare these two schemes with the proposed scheduler. TABLE V reports the results. The lowest error rate obtained by setting $\beta$ to a certain fixed value is 18.6%, achieved at $\beta = 0.6$. The linear scheduler results in the similar error rate of 18.8%. Our $\beta$ scheduler outperforms the two schemes in comparison, showing the effectiveness of the proposed scheduler. Besides effectiveness, the proposed scheduler does not rely on any manual configuration.

TABLE V. ERROR RATES (%) OF WRN28-2 ON CIFAR-100 OBTAINED BY DIFFERENT $\beta$ SCHEDULRING SCHEMES. EACH EXPERIMENT IS RUN THREE TIMES AND THE AVEAGE RESULT IS REPORTED. THE BEST VALUE IS SHOWN IN BOLD AND THE SECOND BEST IS UNDERLINED.

| $\beta$ | 0 | 0.2 | 0.4 | 0.6 | 0.8 | 1 | linear | ours |
|---|---|---|---|---|---|---|---|---|
| Error Rate | 19.7 | 19.0 | 18.9 | 18.6 | 18.9 | 19.9 | 18.8 | **18.4** |

*D. Further Analysis*

We further analyse the impacts of the following configurations on our framework: 1) the number of share blocks and 2) the number of used DA methods.

**Number of shared blocks.** We choose WRN28-2 and train it on CIFAR-100. WRN28-2 consists of a stem, three blocks, and a classification head. we experiment with sharing the stem together with 0, 1, 2, and 3 blocks, respectively. We also include the case in which the stem is not shared, denoted by "-1", where each DNN involved in the proposed method is trained independently without parameter sharing. The results are reported in Fig. 3. It can be observed that our method is less sensitive to configuration of shared blocks, evidenced by the stable performance when the number of shared blocks is set to -1, 0, 1, and 2. When the number of shared blocks is maximised, i.e., 3, the performance much degrades, indicating the importance of keeping some level of independence for the DNN replicates in the proposed method. It is noteworthy that sharing more blocks may reduce the computational costs. Therefore, our experiments use 2 or 3 shared blocks to balance the performance and the computation costs.

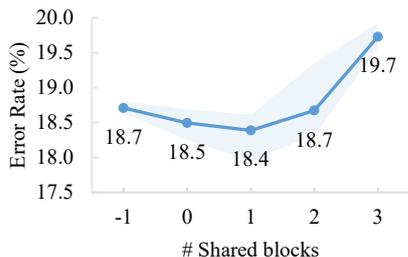

Fig. 3. The error rates of WRN28-2 trained on CIFAR-100 by using the proposed method with the different numbers of shared blocks. Each experiment is run three times. The dots and shadow denote the mean value and the spread of the results over multiple runs.

**Number of data augmentation methods.** We still choose WRN28-2, train it on CIFAR-100 and evaluate the performance of our method in comparison to Baseline 1 when the smaller number of DA methods, i.e., two instead of three, is used. For the fair comparison, we reduce the number of training steps to 1/2 (1/3) for our method armed by two (three) DA methods to ensure that all compared methods consume the same number of training data. We compare our method with Baseline 1 when either three or two DA methods are used, and report the results in TABLE VI. It can be observed that our method consistently outperforms Baseline 1 when either three or two DA methods are used, further verifying the effectiveness of our method. In contrast, using our method with three DA methods outperforms that with two DA methods by 1% no matter which two out of the three are chosen to use.

TABLE VI. Comparison of the Error rates (mean ± std%), averaged over three runs, obtained by WRN28-2 trained on CIFAR-100 via our method and baseline 1 that utilise two and three DA methods, respectively. Each experiment is run three times. The best values are shown in bold.

| DA method | | | Multi-DA method | |
|---|---|---|---|---|
| *RandAugment* | *Mixup* | *CutMix* | *Baseline 1* | *Ours* |
| √ | √ | √ | 19.8 ± 0.3 | **18.4 ± 0.2** |
| √ | √ | | 20.4 ± 0.4 | **19.4 ± 0.2** |
| √ | | √ | 20.1 ± 0.3 | **19.4 ± 0.0** |
| | √ | √ | 20.6 ± 0.2 | **20.1 ± 0.1** |

## V. Conclusion and Future Work

We proposed a multi-DA based DNN training framework that allows making use of arbitrary types of DA methods via KD. To train a specific DNN, our framework replicates a certain portion in the latter part of the DNN into multiple copies, leading to multiple DNNs with shared blocks in their former parts and independent blocks in their latter parts. Each of these DNNs is associated with a unique DA and a newly devised loss with self and mutual parts. This allows the model to comprehensively learn from the data generated by all DA methods and the outputs from all DNNs in an online and adaptive manner. The overall loss, defined as the sum of each DNN's loss, is used for training the DNN. Eventually, one of the DNNs involved in the proposed framework with the best validation performance is chosen for inference. Specifically, the whole training set is randomly split into 80% for training and 20% for validation. We train the DNN on the training set for multiple times using the proposed framework, calculate the average performance of each DNN involved in the framework on the validation set over multiple training runs, and select the best DNN with the best average validation accuracy. Then, we train the DNN on the whole training set, choose the previously selected best DNN, and use it for inference. Experiments of training several representative DNNs on some popular image classification benchmarks demonstrate the superiority of the proposed training method in comparison with several existing single-DA and multi-DA based methods.

Our future work includes but is not limited to evaluation of the performance of the proposed method as the number of the involved DA methods increases, design of certain learnable $\beta$ schedulers, and extensions of the proposed training framework for addressing more different types of machine learning problems and applications [39]–[41] which may benefit from more annotated data.